\documentclass{article}
\usepackage{arxiv}
\usepackage{amsmath,amssymb,mathtools,booktabs,microtype}
\usepackage{hyperref,graphicx}
\usepackage[ruled,vlined,linesnumbered]{algorithm2e}
\usepackage[nameinlink,noabbrev]{cleveref}
\title{Unified Response Geometry for Structured Pruning}
\author{Kaixiang Shu\\Independent Researcher\\\texttt{614729197@qq.com}}
\date{}

\newcommand{\cov}{\operatorname{Cov}}
\newcommand{\corr}{\operatorname{Corr}}
\newcommand{\diag}{\operatorname{diag}}

\begin{document}

\maketitle

\begin{abstract}
Structured pruning is commonly formulated as ranking individual channels, although channel responses can be complementary or cancel through downstream mixing. Motivated by these response interactions, we formulate pruning as the selection of a subset with large joint response capacity, followed by a separate functional realization step. Our unified response geometry maps each candidate set to $M(D,R)=D^{1/2}RD^{1/2}$ and uses its determinant together with Schur-greedy residuals to select non-redundant coordinates. The same construction yields two information-conditioned instances: an unlabeled instance based on activation covariance, and a task-conditioned instance that combines activation and gradient variance for response scale with gradient correlation for complementarity. To convert the selected subset into an executable network, we fold predictable removed responses into successor weights through ridge compensation and recalibrate batch-normalization statistics, without fine-tuning the network. On ImageNet ResNet-50, the unlabeled instance reaches $65.4\%$ and $53.9\%$ Top-1 accuracy at 30\% and 40\% deletion, versus $59.8\%$ and $43.1\%$ for strength-only selection; the task-conditioned instance reaches $67.7\%$ and $56.3\%$ under the same protocol. A six-family screen shows architecture-dependent behavior, with positive relative contrasts in several convolutional and expansion-layer settings and clear boundary cases in windowed attention. These results support response geometry as a conditional principle for structured pruning, with its benefit determined jointly by the observed response and the architecture in which that response is realized.
\end{abstract}

\section{Introduction}
Structured pruning seeks to remove channels, filters or expansion coordinates so that a pretrained network becomes cheaper to execute while retaining its predictive function. A common formulation turns this problem into a ranking of individual coordinates: a scalar score is computed from weights, activations or task sensitivities, and the lowest-scoring coordinates are removed~\cite{lecun1990optimal,han2015learning,han2016deep,liu2017learning,hu2016network,li2017pruning,luo2017thinet,he2017channel,molchanov2019importance}. This formulation is attractive because it gives a simple and controllable pruning rule. It also makes the computational target explicit: the selected coordinates determine the width of a successor layer and therefore the cost of the resulting network. However, the object that is removed is a set of coordinates whose responses are jointly mixed by later layers. The adequacy of an individual score therefore depends on whether the value of a coordinate can be assessed independently of the coordinates retained with it.

Our earlier analysis of convolutional responses showed why this independence assumption can be restrictive. Under downstream linear mixing, channel responses may contain shared components, complementary directions and components that cancel in the readout~\cite{shu2026adjoint}. A channel with a modest marginal amplitude can provide a direction that is difficult to reconstruct from the other retained channels. Conversely, a channel with a large amplitude can be largely redundant with the current set, or its contribution can be offset by another response after mixing. These cases have the same failure mode for pruning: the value of a coordinate is determined by the response space formed with its companions, rather than by its marginal strength alone. The relevant question is consequently not which individual channels are strongest, but which subset preserves the largest amount of non-redundant response capacity at a prescribed width.

Existing criteria expose different parts of this problem. Magnitude and activation statistics describe individual response scale~\cite{lecun1990optimal,liu2017learning,hu2016network,li2017pruning,luo2017thinet,he2017channel}; gradient-based criteria measure sensitivity to a task loss~\cite{molchanov2019importance,lee2019snip}; and geometric or diversity-based methods attempt to avoid selecting similar coordinates~\cite{he2019filter,mariet2016diversity,kulesza2012dpp,lin2020hrank}. These approaches are useful for their respective information regimes, but they generally leave several modeling choices implicit or treat them as separate heuristics: what response is observed, how its scale is assigned, how relations between responses are measured, and how a selected subset is converted into a functioning network. The unresolved problem is therefore to define a subset-level pruning interface that can express both response scale and response relations, accept either unlabeled or task-conditioned observations, and remain separate from the subsequent functional realization of the pruned network.

We address this problem by formulating structured pruning as the selection of a subset with large joint response capacity. For a diagonal scale matrix $D$ and a relationship matrix $R$, we use the unified response geometry
\[
M(D,R)=D^{1/2}RD^{1/2}.
\]
The diagonal term records the scale of each response direction, while the off-diagonal structure records how response directions overlap or complement one another. For a candidate subset, the determinant of its principal matrix measures the volume of the response space jointly retained by that subset. A Schur residual then measures the additional response capacity supplied by a candidate after conditioning on the coordinates already selected. The resulting greedy ordering is therefore defined by a set objective rather than by independent scores, providing a common geometric criterion for deciding which coordinates are worth retaining before the network is edited.

The same interface yields two information-conditioned instances without changing the selection principle. The unlabeled instance obtains both scale and relationships from activation responses, making the criterion applicable when pruning data have no labels. The task-conditioned instance uses activation and gradient variance to define response scale and gradient correlation to describe complementarity relative to the current task. Labels therefore change the response model supplied to the geometry, rather than creating a second pruning principle: the subset objective, determinant/Schur-greedy ordering and realization pipeline remain shared. This separation lets the experiments compare information regimes directly while keeping the combinatorial selection problem fixed.

Selection alone does not specify how the edited network should realize the retained subset. Removing coordinates changes the input seen by the successor layer and can leave normalization statistics calibrated for the unpruned responses. We consequently treat functional realization as a separate step. Predictable components of removed activations are folded into successor weights with a ridge projection, and batch-normalization statistics are recalibrated when such statistics are present. The same realization procedure is applied to every selector, without fine-tuning the backbone or classifier. In this way, the response geometry defines what is retained, while the recovery operator tests whether that geometric choice can be implemented without conflating subset quality with downstream adaptation.

We evaluate the framework in a controlled ImageNet ResNet-50 study, where convolutional channel responses provide the setting directly motivated by the response analysis. In the primary confirmation, the unlabeled instance reaches $65.4\%$ and $53.9\%$ Top-1 accuracy at 30\% and 40\% deletion, compared with $59.8\%$ and $43.1\%$ for the matched strength-only rule; the task-conditioned instance reaches $67.7\%$ and $56.3\%$ under the same protocol. The experiments also compare against classical pruning baselines, isolate the contribution of functional realization, and examine the relationship between response-geometry descriptors and pruning contrasts. We then screen six architecture families to distinguish transfer cases from boundary cases: expansion coordinates in ViT provide a related linear-mixing setting, whereas windowed attention in Swin-T changes the meaning of a local response coordinate. Together, these studies test one unified claim at two levels: response geometry can improve subset selection when the observed responses contain exploitable complementarity, and the size and direction of the resulting contrast depend on how that response is realized by the architecture.

\section{Related Work}
\subsection{Importance- and task-aware pruning}
The dominant formulation of structured pruning assigns an importance score to each removable coordinate and then selects the coordinates with the largest scores under a width or resource budget. Early saliency methods estimated the change in the objective caused by removing parameters~\cite{lecun1990optimal,han2015learning,han2016deep,wen2016structured}. For convolutional networks, Network Slimming uses batch-normalization scale parameters to induce channel sparsity and then removes channels with small learned scales~\cite{liu2017learning,he2017channel,li2017pruning}. Network Trimming instead uses data-dependent neuron responses to identify coordinates that can be removed from a trained model~\cite{hu2016network,luo2017thinet,yu2018nisp,he2018soft,lin2020hrank}. These approaches established the practical value of structured removal: pruning whole channels or filters produces an actual reduction in layer dimensions and can be implemented by standard dense kernels. Structured pruning can also operate within a filter. Pruning Filter in Filter (PFF) learns a Filter Skeleton to prune spatial stripes, each spanning the input channels at a single kernel position~\cite{meng2020pruning}. This exposes pruning granularity as a design choice alongside the selection criterion: PFF learns the spatial shape of individual filters, whereas our framework selects whole channel coordinates according to their joint response geometry.

Importance scores have since been refined in several directions. Layer-adaptive magnitude methods allocate a global sparsity budget across layers instead of applying the same local threshold everywhere~\cite{lee2021layer}. Other criteria estimate the sensitivity of a channel or connection using a first-order Taylor approximation, gradient information, or a one-shot loss perturbation~\cite{molchanov2019importance,lee2019snip,molchanov2019variational}. Activation-weighted criteria extend the same idea to data-dependent responses, including settings where the original training objective is not revisited~\cite{sun2024simple}. These methods are complementary in the information they use, but their common decision variable is still a marginal score for one coordinate. The score can express how large, active or task-sensitive a coordinate is in isolation; it does not by itself specify how the coordinate interacts with the other coordinates that will remain after pruning.

Task-aware pruning is especially useful when labeled examples are available because the score can reflect the task's current decision surface rather than only the magnitude of the internal response. Gradient-based Taylor criteria estimate the first-order change in loss after removing a filter or channel~\cite{molchanov2019importance}, while SNIP and related single-shot approaches use connection sensitivity before training or fine-tuning~\cite{lee2019snip}. Automated policies further learn layer-wise or platform-aware structural decisions, as in MetaPruning, NetAdapt and AMC~\cite{liu2019metapruning,yang2018netadapt,he2018amc}. Unlabeled or activation-based criteria provide a different operating regime: they estimate response statistics from inputs without requiring labels or loss gradients, which is useful for post-training compression and settings where labels are unavailable. In both regimes, however, the information source is commonly used to produce an independent scalar score. A task-sensitive response can therefore receive a high score even when its task effect is already represented by the selected set.

Our framework keeps the scale information represented by these criteria while changing the selection object. The unlabeled instance uses activation covariance to represent response scale and activation relationships, whereas the task-conditioned instance combines activation and gradient variance for scale with gradient correlation for task-conditioned complementarity. Both instances then use the same subset objective and greedy selection rule. The comparison is therefore between information-conditioned response models supplied to one geometric interface, rather than between unrelated scoring heuristics.

Post-training and training-free pruning provide a complementary context for this distinction. SynFlow and GraSP preserve signal flow or gradient flow without access to training examples~\cite{tanaka2020synflow,wang2020grasp}, while lottery-ticket and movement-based approaches study sparse subnetworks through initialization or fine-tuning dynamics~\cite{frankle2019lottery,sanh2020movement}. Recent transformer work extends structural pruning to attention and feed-forward blocks, including compact encoder models and one-shot large-language-model compression~\cite{xia2022structured,frantar2023sparsegpt,ma2023llmpruner}. These methods demonstrate that the available response and adaptation regime strongly affect pruning behavior. Our setting is narrower and more controlled: we study post-training structured coordinate selection with a fixed realization operator, so that the effect of a response-set objective can be separated from training dynamics or extensive retraining.

\subsection{Geometric, diversity, and response-set selection}
Several pruning and compression methods move beyond independent scores by discouraging the selection of similar filters or neurons. Geometric-median pruning removes filters that are close to other filters in a feature space, with the intuition that a nearby filter can provide a substitute for the one removed~\cite{he2019filter}. High-rank feature-map criteria and neuron-importance propagation also use statistics of feature responses or their propagated effects to guide structured removal~\cite{lin2020hrank,yu2018nisp}. Diversity-based compression uses determinantal point processes to favor sets that cover different response directions rather than repeatedly selecting similar elements~\cite{mariet2016diversity,kulesza2012dpp}. These approaches are important precedents for treating the retained set, rather than only its individual members, as an object of optimization.

The motivation for modeling relationships between responses also comes from analyses of downstream mixing in convolutional networks. Our earlier response analysis showed that channel activations can contain shared, complementary and cancelling components when they are combined by later linear operators~\cite{shu2026adjoint}. This observation is compatible with, but not reducible to, a filter-distance criterion: two channels can be close in one representation and still have different effects after the downstream readout, while responses with modest marginal amplitude can provide a direction that is difficult to reconstruct from the current set.

Our response geometry is related to this line of work through its determinant-based set objective, but it makes a different factorization explicit. The matrix $M(D,R)=D^{1/2}RD^{1/2}$ separates the scale of each response from the relationship structure measured across responses. Its principal determinant combines individual response capacity with the complementarity of the selected directions, and its Schur residual gives the incremental capacity of a candidate conditioned on the current set. This provides a fixed-width greedy ordering that can be reused with different response sources. A geometric distance or a diversity kernel can discourage redundancy, but it does not by itself determine which response supplies the scale, how task information should enter, or how the selected subset should be realized in the edited network. The present work turns the response-interaction observation into a pruning interface whose selection principle can be tested through a common functional realization pipeline.

\subsection{Structural surgery and functional recovery}
A ranking rule is not by itself a deployable pruned model. Removing a channel changes the input dimension of the successor layer, and the original successor weights and normalization statistics were calibrated for the unpruned response distribution. Structured pruning systems therefore commonly include a reconstruction, normalization update or subsequent adaptation stage after the structural edit~\cite{hu2016network,molchanov2019importance,he2017channel,he2018soft,yang2018netadapt,he2018amc}. These stages are necessary in practice, but they can also obscure the contribution of the selection criterion if different selectors receive different recovery procedures or if extensive fine-tuning compensates for a weak ranking.

We use functional recovery as a controlled realization step shared by all selectors. A ridge projection predicts the removable response components from the retained ones and folds that prediction into the successor weights; batch-normalization statistics are then recalibrated when the architecture uses running normalization. The recovery operator does not change the selected order and does not fine-tune the backbone or classifier. This separation lets the experiments ask two different questions: which subset has the desired response geometry, and how well can that subset be implemented by the common recovery pipeline? Response geometry defines the subset-level capacity to preserve, while recovery supplies the common realization interface.

\subsection{Position of this work}
The contribution of this paper is a unified response-geometry formulation that connects these lines of work at the level of the retained subset. Relative to marginal importance methods, it adds an explicit representation of cross-coordinate response relations. Relative to geometric and diversity-based selection, it makes the scale--relationship factorization and the response source explicit, and it uses Schur residuals to construct a fixed-width order. Relative to task-aware and unlabeled criteria, it keeps the same selection and realization interface while changing only the observed response model. Relative to reconstruction-based pruning pipelines, it separates the subset objective from the functional recovery applied after structural surgery.

The resulting framework has two final instances in the paper: an unlabeled activation-response instance and a task-conditioned activation/gradient instance. They are not separate principles or historical implementation variants. They are two information-conditioned realizations of one determinant/Schur subset objective, evaluated with one common functional recovery protocol. This organization allows the experiments to compare information regimes directly, trace the response-geometry effect on ResNet-50, and identify where the same principle transfers to expansion coordinates or reaches a boundary in windowed attention.

\section{Method}
\label{sec:method}

\begin{figure}[t]
  \centering
  \includegraphics[width=\linewidth]{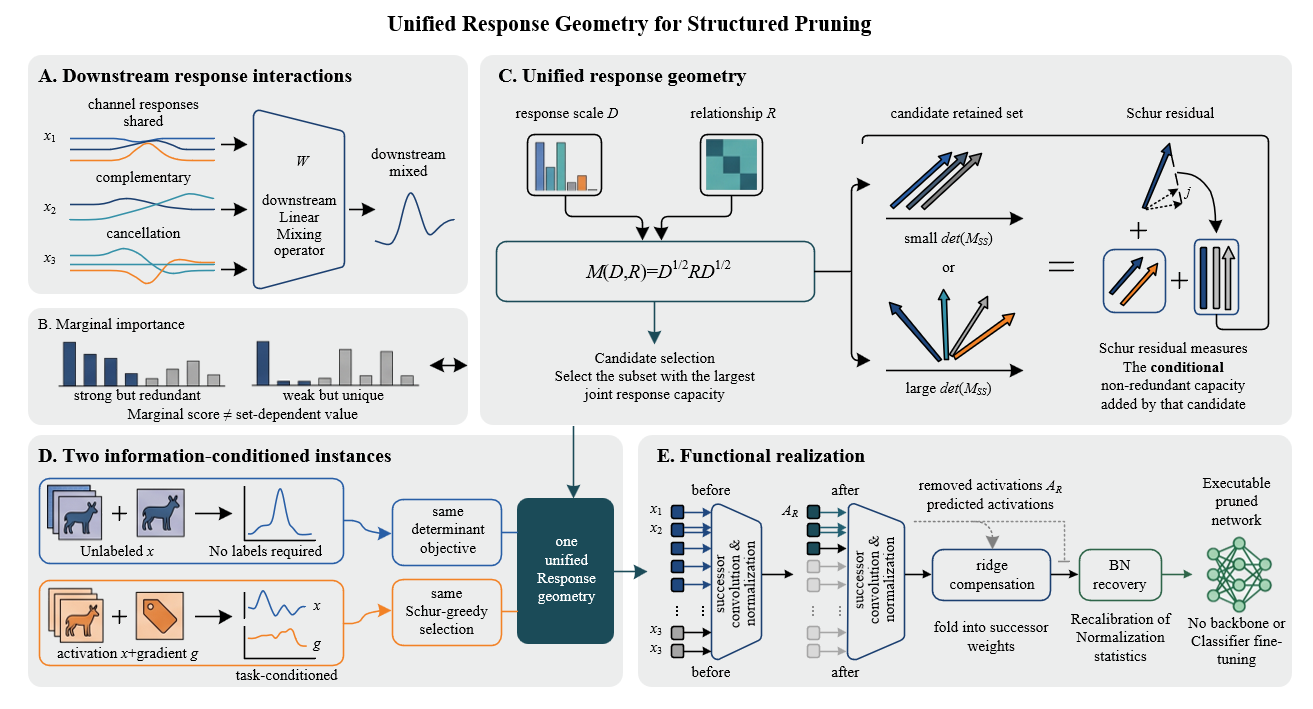}
  \caption{Overview of Unified Response Geometry for Structured Pruning. Downstream responses can be shared, complementary, or cancelling, so marginal importance does not determine the value of a retained set. The unified matrix combines response scale $D$ and relationship structure $R$; determinant-based subset capacity and Schur residuals define the greedy selection. Activation-only and activation--gradient observations instantiate the same objective, after which ridge compensation and BN recovery produce an executable pruned network without backbone or classifier fine-tuning.}
  \label{fig:framework}
\end{figure}

\paragraph{Logic of this section.}
The method has one selection objective and one realization operator, as summarized in Figure~\ref{fig:framework}. We first define the response geometry and its marginal gain, then instantiate the same construction with unlabeled or task-conditioned observations, and finally map the selected coordinates back into a valid network. This ordering makes clear which quantities define a subset and which quantities only restore the network after selection.

\subsection{Pruning objective and response geometry}
\label{sec:geometry}
Consider a pretrained network and a target layer with $C$ removable coordinates. For a prescribed deletion ratio $r$, the selector retains $k=\lfloor(1-r)C\rfloor$ coordinates. Let $S\subseteq\{1,\ldots,C\}$ denote a candidate retained set. The selection objective is local: it asks whether the responses indexed by $S$ span a large and non-redundant response space before the network is edited. It is therefore a criterion for constructing a subset, not a direct theorem about the final task loss.

Let $O\in\mathbb{R}^{N\times C}$ contain centered observations of one response type, with one row per coordinate observation. For a convolutional activation, spatial positions are included as separate rows. The population covariance and its correlation normalization are
\[
 \Sigma_O=\frac{1}{N}O^\top O,\qquad
 R_O=\diag(\Sigma_O)^{-1/2}\Sigma_O\diag(\Sigma_O)^{-1/2},
\]
with an explicit zero-variance convention for degenerate directions. Let $D=\diag(d_1,\ldots,d_C)$ contain the response scale. The unified response-geometry matrix is
\begin{equation}
 M(D,R)=D^{1/2}RD^{1/2}.
 \label{eq:unified}
\end{equation}
The diagonal of $M$ records individual response strength, whereas its off-diagonal entries record relations between response directions. The primary instances use the full relationship matrix; diagonal and interpolated matrices are retained only as prespecified controls in the supplement.

For a positive semidefinite $M$, the combination capacity of $S$ is its squared response volume, with $V(\varnothing)=1$,
\begin{equation}
 V(S)=\det(M_{SS}).
 \label{eq:volume}
\end{equation}
For $j\notin S$, the Schur identity gives
\begin{equation}
 \frac{V(S\cup\{j\})}{V(S)}
 =M_{jj}-M_{jS}M_{SS}^{-1}M_{Sj}.
 \label{eq:schur}
\end{equation}
The right-hand side is the response variance of $j$ that cannot be represented by the selected coordinates. We therefore build an ordering by repeatedly selecting the largest non-negative Schur residual and retaining its first $k$ entries. If numerical rank is exhausted, remaining coordinates are appended in deterministic index order and recorded. For a fixed subset with positive diagonal strengths,
\[
 \log V(S)=\sum_{i\in S}\log d_i+\log\det(R_{SS}),
\]
which makes the scale--complementarity decomposition explicit. The determinant is a selection objective; it is not interpreted as a finite-deletion output-error bound.

\subsection{The two response instances}
\label{sec:instances}
The two proposed rules use identical estimation, ordering, and realization code. They differ in the response used to assign scale and measure complementarity. Let
\(D_x=\diag(\cov(x_1),\ldots,\cov(x_C))\). For the unlabeled instance, the observed response is the activation and
\[
 M_u=D_x^{1/2}\,\corr(x)\,D_x^{1/2}=\cov(x).
\]
For the task-conditioned instance, $g_i$ is the derivative of the current loss with respect to coordinate $i$. We use the task-conditioned scale
\(D_{xg}=\diag(\cov(x_1)\cov(g_1),\ldots,\cov(x_C)\cov(g_C))\) and set
\[
 M_s=D_{xg}^{1/2}\,\corr(g)\,D_{xg}^{1/2}.
\]
Thus the task-conditioned instance uses activation variance for physical amplitude, gradient variance for task sensitivity, and gradient correlation for task-conditioned complementarity. Labels change the observed response and its scale, while the subset size, greedy rule and recovery pipeline remain fixed.
The displayed equalities are population identities. In the finite-sample implementation, the diagonal scale and relationship matrix are estimated from the disjoint banks described below, so their product is a split-sample estimate and need not equal a single-bank empirical covariance entry by entry.

The calibration estimator is deliberately separated from functional adaptation. We use an independent strength bank $\mathcal C_s$ for $D_o$ and relationship bank $\mathcal C_r$ for $R_o$; in the confirmation protocol these contain 2,048 and 4,096 images, respectively. Activations and task responses are collected in evaluation mode with a fixed cross-entropy convention. Counts, means, and centered second moments are accumulated across batches before covariance and correlation normalization. This gives one fixed response matrix per target layer and avoids changing the selector when the adaptation set is changed.

The methodological contribution is the separation of four choices that are often conflated in a pruning score: what response is observed, how individual scale is assigned, how redundancy is measured, and how the selected coordinates are realized in the successor layer. The determinant and Schur identities themselves are standard. The framework makes these choices explicit, keeps the selection and realization interface fixed across the two information regimes, and exposes the response factorization to controlled ablations.

\begin{algorithm}[t]
\caption{Unified response-geometry structured pruning}
\label{alg:unified}
\DontPrintSemicolon
\KwIn{target layer $\ell$ with $C$ removable coordinates; deletion ratio $r$; mode $m\in\{\mathrm{unlabeled},\mathrm{task-conditioned}\}$; strength bank $\mathcal C_s$; relationship bank $\mathcal C_r$; adaptation set $\mathcal A$}
\KwOut{selected set $S$ and a functionally recovered successor layer}
\eIf{$m=\mathrm{unlabeled}$}{collect $x$; set $d_i\leftarrow\cov(x_i)$ and $R\leftarrow\corr(x)$}{collect $x$ and task gradients $g$; set $d_i\leftarrow\cov(x_i)\cov(g_i)$ and $R\leftarrow\corr(g)$}\;
Estimate the moments from $\mathcal C_s$ and $\mathcal C_r$, form $D\leftarrow\diag(d)$, and set $M\leftarrow D^{1/2}RD^{1/2}$\;
$S\leftarrow\varnothing$\tcp*{selected coordinates}
\For{$t\leftarrow1$ \KwTo $k=\lfloor(1-r)C\rfloor$}{
  \eIf{$S=\varnothing$}{compute $\delta_i\leftarrow M_{ii}$ for every $i\notin S$}{compute $\delta_i\leftarrow M_{ii}-M_{iS}M_{SS}^{-1}M_{Si}$ for every $i\notin S$}\;
  $j\leftarrow\arg\max_{i\notin S}\delta_i$; $S\leftarrow S\cup\{j\}$\;
}
Fit ridge predictors from retained to removed activations on $\mathcal A$\;
Fold the predictors into successor weights, reset BN statistics, and recompute them on $\mathcal A$\;
Evaluate the recovered network and report accuracy, teacher KL, and prediction agreement\;
\end{algorithm}

\subsection{Functional realization and recovery}
\label{sec:compensation}
Selection changes the coordinates consumed by the successor layer, so a coordinate order is not yet a deployable pruned network. Let $A_S$ and $A_R$ be the retained and removed activation coordinates collected on an independent adaptation set. We fit a zero-intercept ridge projection
\begin{equation}
 B=\big(\widehat{\mathbb E}[A_S^{\top}A_S]+\rho I\big)^{-1}
    \widehat{\mathbb E}[A_S^{\top}A_R],
 \label{eq:ridge}
\end{equation}
where $\rho$ is proportional to the mean retained second moment. If the successor convolution has input weights $[W_S,W_R]$, substituting $A_R\approx A_SB$ yields
\begin{equation}
 W'_{\mathrm{keep}}=W_S+W_RB^{\top}.
 \label{eq:fold}
\end{equation}
The projection is fitted once per target layer and is never used to alter the selected order. Dependent BN parameters are sliced consistently; residual additions retain their original output width, so the operation removes only the specified input coordinates.

After folding, BN running means and variances are reset and recomputed with cumulative statistics on the adaptation images. We report the compensated network before this step as the \emph{raw} state and the same network after recalibration as the \emph{BN} state. Every selector, including L2 and strength-only baselines, receives the same adaptation images, ridge coefficient, BN procedure, target widths, and evaluation images. Thus the method comparison isolates the response-geometry selection rule, while the raw-to-BN transition exposes how functional recovery contributes to the final result.

\subsection{Assumptions, implementation, and scope}
\label{sec:boundary}
The framework uses response magnitude as an individual scale and correlation as a measure of complementary response directions. The task-conditioned instance adds gradient variance and gradient correlation to describe task sensitivity. These choices are examined with controlled coefficients, numerical checks and the cross-architecture mechanism experiment.

The covariance accumulator costs $O(NC^2)$ time and $O(C^2)$ memory for a layer; constructing a full greedy order costs $O(C^3)$ with rank-one Schur updates. No classifier or backbone weights are trained, and all selection quantities are frozen before evaluation. The primary confirmation therefore tests response geometry under a fixed compensation and BN-recovery pipeline.

\section{Experiments}
\label{sec:experiments}
The experiments test one claim in progressively broader settings: response geometry should change structured selection when the pruning and recovery pipeline is held fixed, while the magnitude and direction of the contrast should depend on the observation model and architecture. We begin with a controlled multi-seed confirmation on ResNet-50, isolate functional realization from subset selection, and then examine transfer and boundary cases across six architecture families. Unless stated otherwise, accuracies are measured after the common recovery step (BN recalibration where BN is present); teacher KL and prediction agreement are fidelity diagnostics.

\subsection{Experimental protocol and controlled comparison}
\label{sec:exp-setup}
The primary confirmation uses one ImageNet-pretrained ResNet-50 checkpoint and 32 internal bottleneck targets, consisting of the paired \texttt{conv1}/\texttt{conv2} layers in the four residual stages. At a target layer with $C$ removable coordinates, every method retains $k=\lfloor(1-r)C\rfloor$ coordinates at deletion ratio $r$. The comparison includes L2 magnitude, a diagonal strength-only rule, and the two final response-geometry instances defined in Section~\ref{sec:instances}. The broader six-family screen adds representative magnitude, activation-statistic, geometric-median and gradient-sensitivity baselines~\cite{liu2017learning,hu2016network,he2019filter,lee2021layer,molchanov2019importance,lee2019snip,sun2024simple,lecun1990optimal,han2015learning,li2017pruning,luo2017thinet,he2017channel,he2018soft,lin2020hrank,liu2019metapruning,liu2019rethinking,yang2018netadapt,he2018amc}. This choice makes the main contrast explicit: L2 and strength-only ignore cross-coordinate response geometry, whereas the two final instances apply one determinant/Schur interface to activation or task-conditioned responses.

The calibration data are partitioned by role. The strength bank contains 2,048 images and determines the diagonal scale; the relationship bank contains 4,096 images and determines the correlation matrix. An independent 3,200-image adaptation bank is used only for ridge compensation and BN recalibration, and a disjoint 2,048-image confirmation bank is used only for evaluation. We repeat the complete calibration and candidate-freezing procedure for three seeds. The checkpoint, target graph and confirmation images are fixed across seeds, so the reported standard deviations quantify calibration variation rather than an additional source of image-level uncertainty.

All selectors are evaluated under the same functional realization. Removed activations are predicted from retained activations with the zero-intercept ridge model, the predictor is folded into the successor convolution, dependent normalization parameters are sliced, and running BN statistics are recomputed on the adaptation bank when BN is present. No classifier or backbone parameter is fine-tuned. This control is essential because a ranking method and a deployable pruned network are different objects: changing the downstream realization can dominate the apparent effect of the ranking. We therefore report the recovered BN endpoint as the primary result and retain raw and teacher-fidelity measurements as diagnostics.

The primary endpoint is ImageNet Top-1 accuracy on the fixed confirmation bank. Teacher KL, centered-logit MSE and prediction agreement are recorded for the same predictions. The seed-level unit of analysis is a paired difference between methods evaluated with the same target structure and confirmation images. This pairing prevents a favorable split or a different recovery state from being mistaken for a selection difference. The protocol and all candidate hashes are frozen before confirmation; the complete moment-estimation and audit details are given in the supplement.

\subsection{Primary response-geometry result on ResNet-50}
\label{sec:exp-main}
Table~\ref{tab:main_accuracy} answers the first question: does adding response geometry to a diagonal strength score change structured-pruning accuracy under the common realization pipeline? At 30\% and 40\% deletion, the unlabeled instance reaches $65.43\pm0.20$ and $53.92\pm0.48$ Top-1, versus $59.81\pm2.17$ and $43.13\pm0.32$ for strength-only selection. The paired contrasts are $+5.62$ and $+10.79$ percentage points, respectively, and all three calibration seeds favor the response-geometry rule. The contrast is larger at the stronger deletion rate, where selecting complementary responses becomes more consequential than preserving only the largest individual variances.

The task-conditioned instance yields the same ordering relative to the diagonal comparator. It reaches $67.66\pm2.12$ and $56.25\pm2.61$ Top-1 at 30\% and 40\% deletion, corresponding to paired contrasts of $+7.85$ and $+13.12$ points; all three seeds favor the task-conditioned rule. Its scale contains both activation and gradient variance, while its relationship term is the gradient correlation. This is the task-conditioned response model used in the confirmation protocol, so the result measures the combined effect of task sensitivity and task-conditioned complementarity rather than the effect of labels alone.

The two information regimes show the same ordering pattern. Activation geometry supplies information beyond a diagonal strength score when labels are unavailable. The task-conditioned instance retains the same volume principle while adding a task response, yielding higher Top-1 accuracy than the diagonal control under the fixed realization pipeline. The comparison is between the two final instances and matched controls; no internal implementation history is treated as a separate method.

\begin{table}[t]
\centering
\caption{Primary BN-recovered Top-1 accuracy (mean $\pm$ s.d. over three calibration seeds).}
\label{tab:main_accuracy}
\begin{tabular}{lcc}
\toprule
Method & 30\% & 40\% \\
\midrule
L2 & 47.69 $\pm$ 1.49 & 31.25 $\pm$ 0.69 \\
Strength-only & 59.81 $\pm$ 2.17 & 43.13 $\pm$ 0.32 \\
Unlabeled instance & 65.43 $\pm$ 0.20 & 53.92 $\pm$ 0.48 \\
Task-conditioned instance & 67.66 $\pm$ 2.12 & 56.25 $\pm$ 2.61 \\
\bottomrule
\end{tabular}
\end{table}

\begin{table}[t]
\centering
\caption{Pre-specified paired Top-1 differences across calibration seeds.}
\label{tab:seed_differences}
\begin{tabular}{llcc}
\toprule
Comparison & Deletion & Mean (pp) & Positive seeds \\
\midrule
Unlabeled instance $-$ strength-only & 30\% & +5.62 & 3/3 \\
Task-conditioned instance $-$ strength-only & 30\% & +7.85 & 3/3 \\
Unlabeled instance $-$ strength-only & 40\% & +10.79 & 3/3 \\
Task-conditioned instance $-$ strength-only & 40\% & +13.12 & 3/3 \\
\bottomrule
\end{tabular}
\end{table}

\begin{figure}[t]
  \centering
  \includegraphics[width=\linewidth]{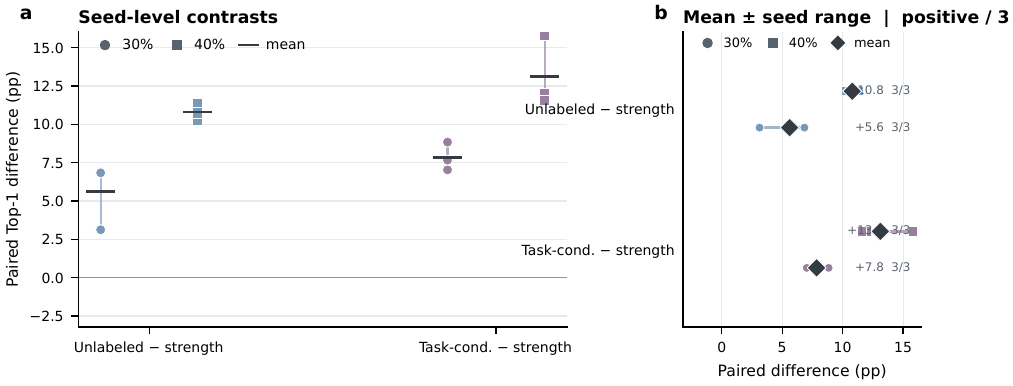}
  \caption{Calibration-seed contrasts for the two final response-geometry instances. Each point is a paired Top-1 difference from strength-only selection on the same confirmation images. Both instances are positive for all three seeds at both deletion rates.}
  \label{fig:seed-gains}
\end{figure}

\subsection{Selection versus functional realization}
\label{sec:exp-recovery}
\begin{figure}[t]
  \centering
  \includegraphics[width=\linewidth]{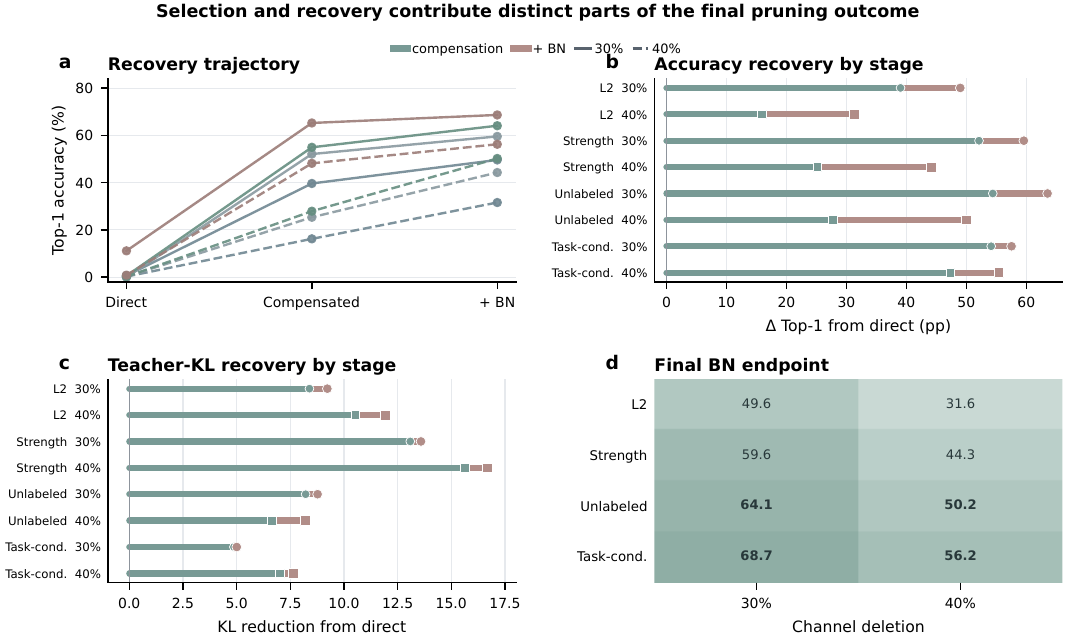}
  \caption{Selection and functional recovery are separate stages in a representative flow diagnostic. Panel a follows direct slicing, ridge compensation and BN recalibration; panels b and c show the corresponding accuracy and teacher-KL increments; panel d reports the final BN endpoint. The same recovery procedure is applied to every selector, so the figure diagnoses the pipeline rather than replacing the multi-seed selection comparison.}
  \label{fig:recovery}
\end{figure}

The second question is whether the response-volume ranking remains useful after the selected coordinates are converted into a valid network. Figure~\ref{fig:recovery} and Table~\ref{tab:recovery_flow} trace the same candidate sets through direct slicing, ridge compensation and BN recalibration. Direct slicing removes coordinates from the successor input while leaving the successor weights and running statistics calibrated for the original coordinates. It is a diagnostic of an incomplete realization, so the flow separates structural editing from the recovery process.

Ridge compensation restores the contribution of removed responses through the retained coordinates. For each target layer, the predictor is fit on the independent adaptation bank and folded into the successor convolution without changing the selected order. The compensated state recovers most of the direct-slice loss across the representative methods in the flow diagnostic. Because the predictor and its coefficient are shared across selectors, this recovery measures the value of a common realization operator rather than a method-specific retraining advantage.

BN recalibration supplies a further, distinct correction. Removing coordinates and folding predictors changes the activation distribution seen by downstream normalization, so the running mean and variance from the unpruned network are no longer appropriate. Recomputing them on the same adaptation bank changes the final endpoint and teacher fidelity. The unpruned control passed through the identical BN procedure quantifies the drift introduced by recalibration itself, preventing the BN increment from being attributed entirely to pruning.

The flow experiment thus supports a separation of roles. The determinant and Schur rule defines the subset geometry, ridge compensation makes that subset executable in the successor layer, and normalization recovery restores the operating statistics where the architecture uses running normalization. The three-seed confirmation table remains the primary selection result; the flow experiment evaluates the volume criterion after realization and exposes how each recovery stage contributes to the final endpoint.

\begin{table}[t]
\centering
\caption{Selection versus functional realization in the representative flow diagnostic.}
\label{tab:recovery_flow}
\begin{tabular}{lccccc}
\toprule
Method & Deletion & Direct & Compensated & +BN & Change to +BN (pp) \\
\midrule
L2 & 30\% & 0.6 & 39.6 & 49.6 & +49.0 \\
L2 & 40\% & 0.2 & 16.2 & 31.6 & +31.3 \\
Strength-only & 30\% & 0.0 & 52.1 & 59.6 & +59.6 \\
Strength-only & 40\% & 0.2 & 25.3 & 44.3 & +44.1 \\
Unlabeled instance & 30\% & 0.6 & 55.0 & 64.1 & +63.5 \\
Unlabeled instance & 40\% & 0.2 & 27.9 & 50.2 & +50.0 \\
Task-conditioned instance & 30\% & 11.1 & 65.3 & 68.7 & +57.5 \\
Task-conditioned instance & 40\% & 0.9 & 48.2 & 56.2 & +55.4 \\
\bottomrule
\end{tabular}
\end{table}

\subsection{Cross-architecture behavior and mechanism boundary}
\label{sec:exp-cross}
The final question is how far the response-geometry construction transfers beyond the theory-native CNN setting. We screen six distinct families—ResNet-50, ConvNeXt-B, EfficientNet-B3, MobileNetV2, ViT-S/16 and Swin-T~\cite{he2016deep,liu2022convnet,tan2019efficientnet,sandler2018mobilenetv2,dosovitskiy2021image,liu2021swin}—using the same selector, structural surgery and normalization-recovery pipeline, with architecture-specific dependency handling where needed. Each model contributes 18 selection entries at five deletion rates, giving 540 frozen rows. The release uses the final unlabeled and task-conditioned instances across all six models, with the task-conditioned rows regenerated from the final $D_{xg}$ implementation. It tests transfer and maps the operating range; the multi-seed ResNet-50 experiment supplies the replicated formula-level confirmation.

At 40\% deletion, the unlabeled-minus-L2 Top-1 contrasts are 15.2, 33.1, 4.3, $-2.7$, 24.6 and $-10.7$ points in the model order above. The corresponding task-conditioned-minus-Fisher contrasts are 8.3, $-2.0$, 18.8, 9.8, 14.0 and 4.7 points. The unlabeled instance has the highest reported accuracy in the ResNet-50 and ViT-S/16 cases and at the stronger EfficientNet-B3 budgets, whereas APoZ~\cite{hu2016network}, LAMP~\cite{lee2021layer} or magnitude rules are higher in parts of ConvNeXt-B, MobileNetV2 and Swin-T. The task-conditioned instance has the highest reported accuracy in several ResNet-50, EfficientNet-B3 and MobileNetV2 budgets, while Taylor/Fisher-gate~\cite{molchanov2019importance}, OBD~\cite{lecun1990optimal} or Wanda~\cite{sun2024simple} are higher in parts of ConvNeXt-B, ViT-S/16 and Swin-T. The comparison therefore shows architecture-dependent performance rather than uniform dominance.

The architecture pattern follows the scope of the Project1 motivation. Project1 was derived for CNN channel responses with spatially repeated observations, downstream linear mixing and shared cancellation structure; CNN bottleneck and expansion layers therefore form the theory-native domain of the construction. ViT-S/16 provides a transfer case because its MLP hidden units are also linear expansion coordinates, and the unlabeled instance is competitive there, but the Project1 GAP and pixel-space argument does not directly cover ViT. Swin-T is a boundary case: windowed token mixing, LayerNorm and residual paths change the functional meaning of a local response coordinate, and LAMP/Wanda have higher accuracy in the reported budgets. This gives an architecture-conditioned interpretation: CNN is the native domain, ViT is an empirically transferable extension, and Swin-T defines a boundary for the present response model.

We next test that interpretation directly. Before pruning, we measure shared low-rank energy, a fixed downstream-readout cancellation residual and normalized effective covariance rank for each model, aggregate them over prunable layers, and relate every descriptor to matched unlabeled--L2 and task-conditioned--Fisher contrasts. The cancellation residual has the clearest directional association with the unlabeled contrast at 40\% deletion ($\rho=0.89$, exact permutation $p=0.035$, $n=6$). Low-rank energy is not monotone, and task-conditioned associations are mixed; the complete descriptor matrix and all 24 correlations are provided in the supplement. Thus the mechanism experiment supports a measurable cancellation component while rejecting a single-scalar explanation of all architecture behavior. This test is motivated by the superposition and cancellation observations in Project1~\cite{shu2026adjoint}.

Together, the confirmation and screen show that response geometry changes selection outcomes under the fixed realization protocol used for each study. The replicated ResNet-50 confirmation establishes the main result, while the fixed-budget screen maps its transfer range and architecture-dependent contrasts.

\begin{figure}[t]
  \centering
  \includegraphics[width=\linewidth]{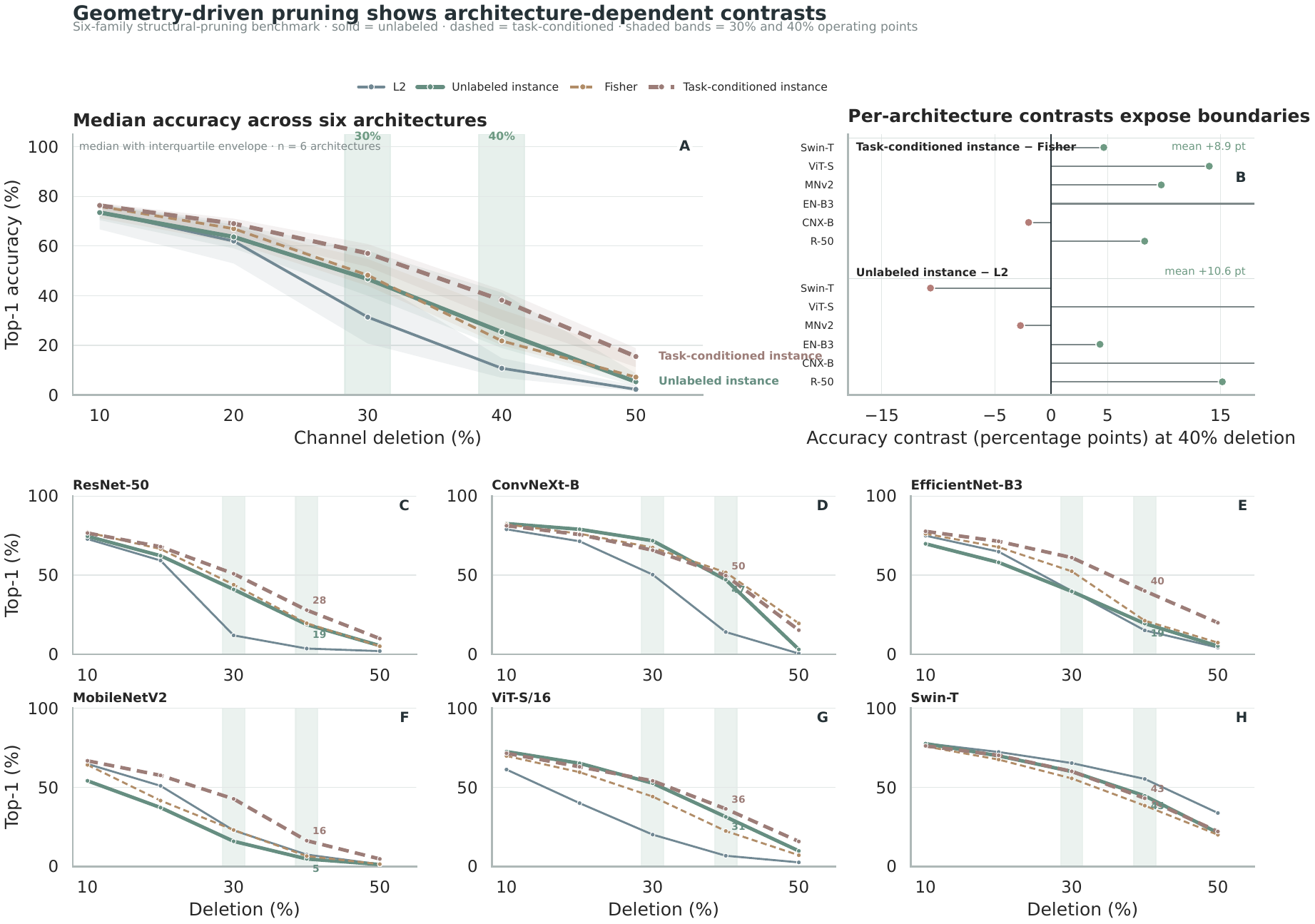}
  \caption{Six-family cross-architecture screening. Panel A summarizes the median accuracy trajectory and interquartile spread; panel B reports paired 40\% contrasts against regime-matched comparators; panels C--H show each architecture. Negative contrasts remain visible to define the operating boundary.}
  \label{fig:multimodel}
\end{figure}
\begin{table*}[t]
\centering\small
\caption{Six-family structural-pruning screening. Entries are Top-1 accuracy (\%). “Best U/L” is the best method within the unlabeled or label/gradient comparator family at the same deletion rate; the two final response-geometry instances are shown separately so architecture boundaries remain visible.}
\label{tab:multimodel-main}
\resizebox{\textwidth}{!}{%
\begin{tabular}{l*{12}{c}}
\toprule
Model & \multicolumn{4}{c}{30\%} & \multicolumn{4}{c}{40\%} & \multicolumn{4}{c}{50\%} \\
& Unlabeled & best U & Task-cond. & best L & Unlabeled & best U & Task-cond. & best L & Unlabeled & best U & Task-cond. & best L \\
\midrule
ResNet-50 & 40.9 & 40.9 (Unlabeled instance) & 50.8 & 50.8 (Task-conditioned instance) & 18.8 & 18.8 (Unlabeled instance) & 27.9 & 27.9 (Task-conditioned instance) & 5.4 & 5.4 (Unlabeled instance) & 9.9 & 9.9 (Task-conditioned instance) \\
ConvNeXt-B & 71.7 & 72.5 (APoZ) & 65.7 & 72.5 (Taylor-gate) & 47.1 & 47.2 (APoZ) & 49.7 & 56.0 (Fisher-gate) & 3.0 & 3.8 (APoZ) & 15.2 & 24.0 (OBD) \\
EfficientNet-B3 & 39.7 & 41.0 (FPGM) & 61.0 & 61.0 (Task-conditioned instance) & 19.4 & 19.4 (Unlabeled instance) & 40.0 & 40.0 (Task-conditioned instance) & 5.2 & 5.2 (Unlabeled instance) & 19.9 & 19.9 (Task-conditioned instance) \\
MobileNetV2 & 15.8 & 23.5 (L1) & 42.6 & 42.6 (Task-conditioned instance) & 4.7 & 7.4 (LAMP) & 16.2 & 16.2 (Task-conditioned instance) & 1.2 & 1.3 (LAMP) & 4.7 & 4.7 (Task-conditioned instance) \\
ViT-S/16 & 52.5 & 52.5 (Unlabeled instance) & 54.0 & 54.0 (Task-conditioned instance) & 31.3 & 31.3 (Unlabeled instance) & 36.3 & 42.5 (Fisher-gate) & 9.8 & 9.8 (Unlabeled instance) & 15.8 & 21.3 (Fisher-gate) \\
Swin-T & 59.9 & 65.2 (LAMP) & 60.0 & 65.0 (Wanda) & 44.5 & 55.2 (LAMP) & 43.0 & 50.6 (Wanda) & 21.2 & 33.7 (LAMP) & 22.0 & 25.4 (Wanda) \\
\bottomrule
\end{tabular}
}%
\end{table*}

\begin{figure}[t]
  \centering
  \includegraphics[width=\linewidth]{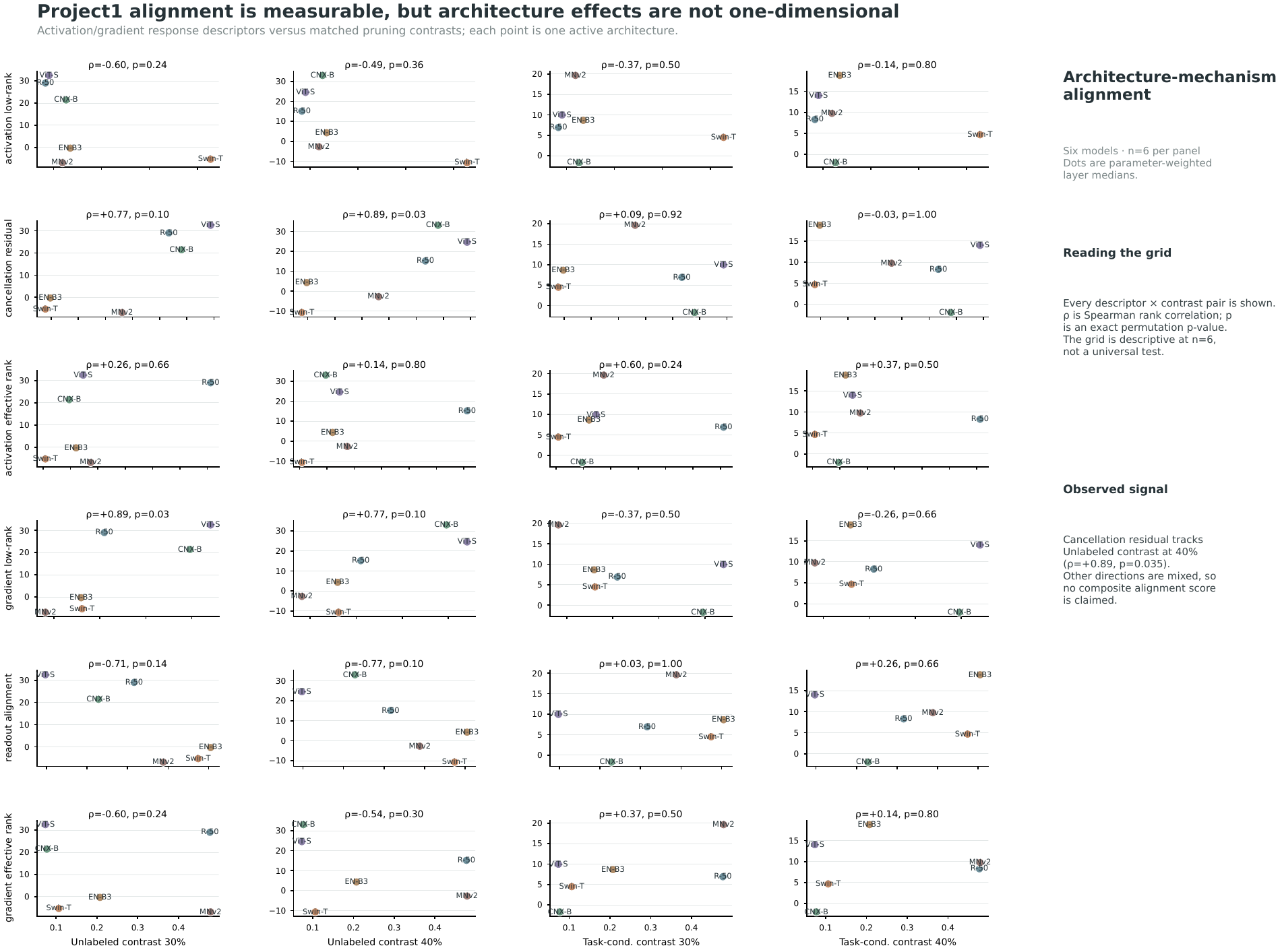}
  \caption{Architecture--mechanism alignment. Descriptor--contrast pairs are shown for all six architectures. The strongest directional association is between cancellation residual and the unlabeled contrast at 40\%, while low-rank and task-conditioned relationships remain mixed.}
  \label{fig:architecture-mechanism}
\end{figure}

\section{Discussion}
\label{sec:discussion}
The central implication of the results is that structured pruning can be viewed as a response-modeling problem rather than only a ranking problem. The relevant question is which response directions remain jointly useful at the target width, not which coordinates have the largest marginal scores. In the ResNet-50 confirmation, both information-conditioned instances exceed the strength-only comparator under the same recovery procedure across all calibration seeds. For the unlabeled instance, activation relationships provide selection information beyond individual response variance, which is particularly useful when task feedback is unavailable. The task-conditioned instance applies the same principle to a response model that incorporates task sensitivity: its activation/gradient scale and gradient correlation allow task information to affect both individual strength and complementarity. The common construction therefore connects the two information regimes through one shared question: how much additional response capacity does a coordinate provide alongside those already retained?

The recovery experiment gives this geometric statement an operational meaning. The volume objective favors responses that supply distinct directions at the selected width, while the ridge projection uses retained activations to reproduce predictable components of the removed responses. These operations act on different aspects of compression: selection determines the available response coordinates, and compensation adjusts the successor to use them. BN recalibration then aligns running statistics with the edited network. The progression from direct slicing through compensation to recalibration in Figure~\ref{fig:recovery} illustrates why final accuracy depends on the realization process as well as on the retained set. Keeping the recovery procedure fixed across selectors makes the comparison informative about the selection criterion within that pipeline. This separates the geometric hypothesis from its functional consequence: response capacity defines the property being retained, while recovered accuracy and teacher fidelity test how that property survives in the edited network.

The cross-architecture results locate this principle within the response structures that motivated it. Convolutional channels provide spatially repeated observations followed by linear mixing, which connects the CNN setting to the earlier analysis of superposition and cancellation. The competitive unlabeled results on ViT-S/16 extend the empirical evidence to MLP expansion coordinates with a linear successor. Swin-T, together with mixed results in some ConvNeXt-B and MobileNetV2 budgets, shows that a shared architecture label alone does not determine the best selector. The coordinate being removed and the surrounding mixing and normalization operations matter for interpreting its response geometry. The mechanism experiment provides a measurable link to this interpretation: cancellation residual has the clearest association with the unlabeled contrast at stronger deletion, whereas shared low-rank energy and effective rank do not consistently explain the ordering across models. This pattern is consistent with a role for downstream response interactions, while the mixed task-conditioned associations indicate that task sensitivity introduces additional structure. The resulting scope is therefore architectural and conditional: the response geometry is most directly motivated for CNN channels, can transfer to related expansion coordinates, and requires a different response model when the meaning of a local coordinate changes.

The formulation also identifies concrete directions for extending its practical range. Estimating full response relationships requires quadratic memory in layer width, and constructing a complete Schur-greedy order has cubic cost. Low-rank or blockwise approximations are therefore natural directions for wide layers, with their quality assessed by the resulting coordinate selections and recovered predictions. The task-conditioned instance additionally requires labels and gradients for the chosen loss, making the calibration task part of the response definition. A complementary modeling question is how to choose the observation unit when normalization or token mixing couples coordinates beyond a single channel. Finally, the link between the volume objective and reconstruction under a fixed downstream readout remains a theoretical question: the present evidence establishes usefulness through the recovered network. These questions preserve the framework's central separation between response modeling, subset selection and functional realization, while identifying where further analysis can make their connection more precise.

\section{Conclusion}
\label{sec:conclusion}
We presented Unified Response Geometry for Structured Pruning, a framework that separates response scale from response relationships and selects retained coordinates through a determinant objective and Schur-greedy ordering. Activation and task-conditioned observations yield two instances of the same construction, followed by functional realization through ridge compensation and applicable BN recalibration. The controlled ImageNet ResNet-50 experiments show that both instances retain higher accuracy than strength-only selection without network fine-tuning. The six-family evaluation finds transfer in additional convolutional and expansion-layer settings while identifying architecture-dependent boundaries, and the mechanism analysis connects part of this variation to downstream response interactions. Together, these findings support a conditional organizing principle for structured pruning: when the observed coordinates contain exploitable complementary responses, the retained set should be selected by its joint response capacity and judged by the function realized by the edited network.

\clearpage
\appendix
\section*{Supplementary Material}
\addcontentsline{toc}{section}{Supplementary Material}

\section{Response estimators and calibration protocol}
\label{sec:stats}
The supplement provides the implementation detail needed to reproduce the response matrices and to distinguish calibration uncertainty from functional adaptation. The main paper defines the common matrix and selection rule; this section fixes the observation units, moment estimator, gradient convention, and the two independent calibration banks used by the confirmation protocol.

For a convolutional activation $x\in\mathbb{R}^{B\times C\times H\times W}$, each spatial position is treated as one coordinate-response observation. For wide layers, the implementation applies a deterministic evenly spaced cap of 4,096 rows per physical batch to keep the covariance construction tractable; otherwise all available spatial rows are used. The task-gradient tensor uses the same row convention and cap after differentiating the saved activation under the fixed cross-entropy convention. This produces a covariance of task-response variation.

The estimator maintains the sample count, mean and centered second-moment matrix for every target layer. When two accumulators are merged, the between-mean correction is applied explicitly,
\[
S_{12}=S_1+S_2+\frac{n_1n_2}{n_1+n_2}(\mu_2-\mu_1)(\mu_2-\mu_1)^\top.
\]
The covariance is the population convention $S/(n_1+n_2)$ used by the implementation. Correlations are normalized from the covariance diagonal; zero-variance directions receive zero off-diagonal correlations and a unit diagonal only when that direction is defined.

Let $\Sigma_x$ and $\Sigma_g$ denote the activation and gradient covariances. The unlabeled instance uses $d_i^{(u)}=(\Sigma_x)_{ii}$ and $R_x=\operatorname{Corr}(x)$. The task-conditioned instance uses $d_i^{(s)}=(\Sigma_x)_{ii}(\Sigma_g)_{ii}$ and $R_g=\operatorname{Corr}(g)$, giving
\[
M_u=D_x^{1/2}R_xD_x^{1/2},\qquad
M_s=D_{xg}^{1/2}R_gD_{xg}^{1/2}.
\]
The gradient variance in $D_{xg}$ supplies task-sensitive response scale, while the gradient correlation supplies the relationship term. This factorization is fixed before confirmation.
These equations describe the population construction. Because the implementation estimates the diagonal scale and relationship matrix from disjoint calibration banks, the finite-sample product is a split-sample estimate and is not required to match a covariance computed from one bank entry by entry.

The strength bank $\mathcal C_s$ and relationship bank $\mathcal C_r$ are disjoint. In the confirmation protocol they contain 2,048 and 4,096 images, respectively. The adaptation bank contains 3,200 images and the confirmation bank contains 2,048 held-out validation images. Candidate sets are frozen before confirmation and all methods receive the same target widths, adaptation images, compensation coefficient, normalization-recovery procedure (BN where present) and confirmation images. This separation prevents the recovery set or evaluation outcome from changing the response matrix.

\section{Determinant geometry and controlled design choices}
\label{sec:derivation}
The selection objective is the squared volume of the response ellipsoid generated by a candidate subset, consistent with determinant-based diversity selection~\cite{supp:mariet2016diversity}. If $M$ is positive semidefinite and $S$ is a selected index set, the volume factor is $V(S)=\det(M_{SS})$. For a new coordinate $j$, with $V(\varnothing)=1$, the Schur identity gives
\[
\det(M_{S\cup\{j\},S\cup\{j\}})=\det(M_{SS})\left(M_{jj}-M_{jS}M_{SS}^{-1}M_{Sj}\right).
\]
The parenthesized term is the residual response variance after projection onto the selected coordinates. The greedy solver therefore appends the largest non-negative residual at each step. If numerical rank is exhausted, the remaining indices are appended in deterministic index order and logged.

For the primary rule, a fixed subset with positive diagonal strengths separates scale and complementarity:
\[
\log\det(M_{SS})=\sum_{i\in S}\log d_i+\log\det(R_{SS}).
\]
For the control study only, we replace $R$ by $R_\lambda=(1-\lambda)I+\lambda R$ to interpolate between diagonal strength and full geometry. The identity separates the two contributions without changing the primary selector.

The controlled study varies the relationship contribution while holding the strength bank fixed and changing relationship-bank size from 1,024 to 4,096 images. The unlabeled $\lambda=0.5$ rule has paired contrasts of $+1.42$--$+1.69$ points relative to its matched diagonal across the four budget/sample-size conditions, with all 12 seed-level differences positive. The full relationship is retained as the primary setting because it is fixed before confirmation and does not depend on accuracy-based tuning; the control results and relationship-repeatability analysis are reported for completeness.

The controlled interpretation is therefore deliberately limited. Activation magnitude supplies physical response scale, gradient variance adds task sensitivity in the task-conditioned instance, and correlation supplies a redundancy relation. The experiments test this factorization under the fixed intervention pipeline.

\section{Functional realization and implementation audit}
\label{sec:comp-detail}
A selected subset becomes deployable only after the coordinates removed from a layer are accounted for in its successor. Let $a_S$ and $a_R$ be retained and removed activation coordinates on the adaptation bank. We fit a zero-intercept ridge predictor,
\[
B=\left(\widehat{\mathbb E}[a_S^\top a_S]+\rho I\right)^{-1}
  \widehat{\mathbb E}[a_S^\top a_R],
\]
where $\rho$ is proportional to the mean retained second moment. If the successor convolution has input weights $[W_S,W_R]$, replacing $a_R$ by $a_SB$ gives the folded retained weight $W_S+W_RB^\top$. The predictor is fit once per target layer and never changes the selected order.

The structural edit slices dependent normalization parameters and preserves the output width of residual additions. For the bottleneck targets, the first successor convolution receives the folded contribution from removed coordinates, while the second target removes its output coordinates and slices the following convolution consistently. The direct-slice state ($t=0$) and compensated state ($t=1$) are both retained in the flow diagnostic. No classifier or backbone parameter is trained.

After folding, BN running means and variances are reset and recomputed cumulatively on the adaptation bank. The unpruned network passes through the same BN procedure as a control, because recalibration can itself alter teacher agreement. The primary endpoint is the compensated network after this BN step; raw Top-1, teacher KL, centered-logit MSE and prediction agreement expose the intermediate changes without replacing the primary endpoint.

The implementation audit checks compressed and dense masked forwards before scoring, exact target widths, unpruned identity before BN recovery, deterministic rank-exhaustion behavior, cached image and logit hashes, checkpoint hashes and candidate manifests. The wide-layer GPU Schur solver used by the six-model release agrees with the CPU reference to approximately $1.5\times10^{-7}$ in covariance/ordering checks. Together, these checks establish implementation consistency for the reported runs.

\section{Confirmation statistics and secondary endpoints}
\label{sec:protocol}
The formal ResNet-50 confirmation uses one ImageNet-pretrained checkpoint and 32 internal bottleneck \texttt{conv1}/\texttt{conv2} targets. For each of three calibration seeds, the selection, relationship, adaptation and confirmation banks are separate and contain 2,048, 4,096, 3,200 and 2,048 images. The two final instances and the L2 and strength-only controls use 30\% and 40\% coordinate deletion with $k=\lfloor(1-r)C\rfloor$.

The candidate structures are frozen before evaluation. Identical structures share one confirmation forward, while every method receives the same compensation and normalization-recovery procedure (BN where present). The primary endpoint is accuracy after the applicable recovery step, with BN-recovered Top-1 as the CNN endpoint. Teacher KL measures divergence from the unpruned teacher distribution, centered-logit MSE measures logit displacement after removing the teacher mean, and agreement measures the fraction of examples with identical top-1 predictions. These metrics answer different questions and are therefore reported separately rather than collapsed into one score.

\begin{table}[t]
\centering
\caption{Secondary BN metrics averaged over calibration seeds.}
\label{tab:secondary_metrics}
\begin{tabular}{llccc}
\toprule
Method & Deletion & Top-1 (\%) & KL $\downarrow$ & Agreement (\%) \\
\midrule
L2 & 30\% & 47.69 & 2.139 & 51.35 \\
L2 & 40\% & 31.25 & 3.272 & 33.06 \\
Strength-only & 30\% & 59.81 & 1.253 & 64.58 \\
Strength-only & 40\% & 43.13 & 2.369 & 46.13 \\
Unlabeled instance & 30\% & 65.43 & 0.890 & 71.50 \\
Unlabeled instance & 40\% & 53.92 & 1.536 & 57.80 \\
Task-conditioned instance & 30\% & 67.66 & 0.757 & 73.97 \\
Task-conditioned instance & 40\% & 56.25 & 1.384 & 60.53 \\
\bottomrule
\end{tabular}
\end{table}

\begin{table}[t]
\centering
\caption{Pre-specified paired Top-1 differences across calibration seeds.}
\label{tab:supp_seed_differences}
\begin{tabular}{llcc}
\toprule
Comparison & Deletion & Mean (pp) & Positive seeds \\
\midrule
Unlabeled instance $-$ strength-only & 30\% & +5.62 & 3/3 \\
Task-conditioned instance $-$ strength-only & 30\% & +7.85 & 3/3 \\
Unlabeled instance $-$ strength-only & 40\% & +10.79 & 3/3 \\
Task-conditioned instance $-$ strength-only & 40\% & +13.12 & 3/3 \\
\bottomrule
\end{tabular}
\end{table}

Table~\ref{tab:secondary_metrics} gives all secondary endpoints for the two final instances and their controls. Table~\ref{tab:supp_seed_differences} gives paired seed differences for the pre-specified contrasts. Seed-level differences are reported directly; any image bootstrap interval conditions on a fixed calibration seed and is not treated as an independent replication. All paired contrasts for the two final instances relative to strength-only selection are positive for the three calibration seeds.

\section{Extended screening and mechanism analysis}
\label{sec:extended}
The six-family release contains ResNet-50, ConvNeXt-B, EfficientNet-B3, MobileNetV2, ViT-S/16 and Swin-T, with 18 selection entries at each of five deletion rates. All models use the common fixed-budget screening protocol recorded in the release manifest and the final unlabeled and task-conditioned instances, with architecture-specific dependency handling where needed. The task-conditioned rows were regenerated with the final $D_{xg}$ implementation. The complete tables below report every configured method at the two deletion rates emphasized in the main paper; the release files retain all five rates. The tables expose transfer trends and architecture boundaries, while the replicated ResNet-50 experiment remains the multi-seed formula-level confirmation.

\begin{table*}[t]
\centering\scriptsize
\caption{Complete unlabeled comparison at 30\% deletion. Top-1 accuracy (\%).}
\label{tab:multimodel_30_u}
\begin{tabular}{l*{6}{r}}
\toprule
Method & ResNet-50 & ConvNeXt-B & EfficientNet-B3 & MobileNetV2 & ViT-S/16 & Swin-T \\
\midrule
Random & 21.7 & 66.3 & 31.9 & 8.9 & 41.6 & 60.1 \\
L1 & 5.6 & 24.5 & 36.6 & 23.5 & 13.2 & 63.0 \\
L2 & 11.9 & 50.3 & 40.0 & 22.7 & 20.1 & 65.2 \\
BN-$\gamma$ & 6.0 & 24.5 & 33.2 & 8.2 & 13.2 & 63.0 \\
APoZ & 22.9 & 72.5 & 39.8 & 4.7 & 45.6 & 63.0 \\
LAMP & 11.9 & 50.3 & 40.0 & 22.7 & 20.1 & 65.2 \\
FPGM & 10.7 & 51.2 & 41.0 & 20.0 & 27.9 & 62.7 \\
Unlabeled instance & 40.9 & 71.7 & 39.7 & 15.8 & 52.5 & 59.9 \\
\bottomrule
\end{tabular}
\end{table*}

\begin{table*}[t]
\centering\scriptsize
\caption{Complete label/gradient comparator family at 30\% deletion. Top-1 accuracy (\%).}
\label{tab:multimodel_30_l}
\begin{tabular}{l*{6}{r}}
\toprule
Method & ResNet-50 & ConvNeXt-B & EfficientNet-B3 & MobileNetV2 & ViT-S/16 & Swin-T \\
\midrule
Taylor & 37.0 & 67.4 & 53.8 & 32.5 & 43.5 & 41.4 \\
GradNorm & 44.9 & 68.3 & 44.3 & 18.8 & 40.7 & 51.6 \\
Fisher & 43.9 & 67.4 & 52.4 & 23.0 & 44.1 & 55.6 \\
Taylor-w & 41.7 & 68.5 & 47.8 & 24.3 & 49.9 & 54.0 \\
Taylor-gate & 30.9 & 72.5 & 50.0 & 32.6 & 47.9 & 52.0 \\
Fisher-gate & 29.9 & 69.7 & 51.6 & 28.7 & 53.3 & 58.7 \\
SNIP & 41.7 & 68.5 & 47.8 & 24.3 & 49.9 & 54.0 \\
Wanda & 30.4 & 24.9 & 40.5 & 22.1 & 10.1 & 65.0 \\
OBD & 44.8 & 67.3 & 60.8 & 37.0 & 46.6 & 55.5 \\
Task-conditioned instance & 50.8 & 65.7 & 61.0 & 42.6 & 54.0 & 60.0 \\
\bottomrule
\end{tabular}
\end{table*}

\begin{table*}[t]
\centering\scriptsize
\caption{Complete unlabeled comparison at 40\% deletion. Top-1 accuracy (\%).}
\label{tab:multimodel_40_u}
\begin{tabular}{l*{6}{r}}
\toprule
Method & ResNet-50 & ConvNeXt-B & EfficientNet-B3 & MobileNetV2 & ViT-S/16 & Swin-T \\
\midrule
Random & 4.2 & 39.2 & 13.5 & 1.5 & 16.1 & 45.9 \\
L1 & 0.9 & 2.0 & 9.2 & 3.0 & 4.4 & 51.8 \\
L2 & 3.6 & 14.0 & 15.1 & 7.4 & 6.7 & 55.2 \\
BN-$\gamma$ & 0.4 & 2.0 & 11.8 & 1.7 & 4.4 & 51.8 \\
APoZ & 5.5 & 47.2 & 16.8 & 1.4 & 22.1 & 52.2 \\
LAMP & 3.6 & 14.0 & 15.1 & 7.4 & 6.7 & 55.2 \\
FPGM & 3.3 & 14.2 & 14.5 & 4.1 & 8.2 & 53.3 \\
Unlabeled instance & 18.8 & 47.1 & 19.4 & 4.7 & 31.3 & 44.5 \\
\bottomrule
\end{tabular}
\end{table*}

\begin{table*}[t]
\centering\scriptsize
\caption{Complete label/gradient comparator family at 40\% deletion. Top-1 accuracy (\%).}
\label{tab:multimodel_40_l}
\begin{tabular}{l*{6}{r}}
\toprule
Method & ResNet-50 & ConvNeXt-B & EfficientNet-B3 & MobileNetV2 & ViT-S/16 & Swin-T \\
\midrule
Taylor & 14.2 & 46.6 & 31.2 & 9.0 & 24.9 & 19.4 \\
GradNorm & 24.2 & 51.4 & 19.0 & 6.1 & 18.3 & 30.0 \\
Fisher & 19.6 & 51.7 & 21.2 & 6.5 & 22.3 & 38.4 \\
Taylor-w & 18.5 & 52.1 & 14.9 & 4.0 & 30.2 & 34.8 \\
Taylor-gate & 8.2 & 54.5 & 23.7 & 5.8 & 32.2 & 31.4 \\
Fisher-gate & 7.5 & 56.0 & 22.9 & 5.1 & 42.5 & 42.1 \\
SNIP & 18.5 & 52.1 & 14.9 & 4.0 & 30.2 & 34.8 \\
Wanda & 8.0 & 1.1 & 10.6 & 4.9 & 5.0 & 50.6 \\
OBD & 21.1 & 52.8 & 36.7 & 11.5 & 27.0 & 37.5 \\
Task-conditioned instance & 27.9 & 49.7 & 40.0 & 16.2 & 36.3 & 43.0 \\
\bottomrule
\end{tabular}
\end{table*}

At 40\% deletion, the final task-conditioned-minus-Fisher~\cite{supp:molchanov2019importance} Top-1 contrasts are 8.3, $-2.0$, 18.8, 9.8, 14.0 and 4.7 percentage points for ResNet-50, ConvNeXt-B, EfficientNet-B3, MobileNetV2, ViT-S/16 and Swin-T, respectively. The negative ConvNeXt-B value is retained as an architecture boundary; these contrasts come from fixed-budget screening rather than independent-seed estimates.

The architecture pattern follows the response structures motivating the framework: CNN bottleneck and expansion layers are the theory-native setting because Project1 uses spatially repeated channel responses, downstream linear mixing and cancellation; ViT MLP hidden units are a transferable expansion-layer case not directly covered by the Project1 GAP/pixel-space argument; and Swin-T is a boundary where windowed token mixing, LayerNorm and residual paths alter coordinate function.

The mechanism test asks whether the Project1 interpretation predicts the screening pattern~\cite{supp:shu2026adjoint}. For each model, we aggregate shared low-rank energy, cancellation residual and normalized effective covariance rank over prunable layers with parameter-weighted medians. The contrast endpoints are read from the frozen release, and no descriptor, layer or deletion rate is selected from accuracy. Cancellation residual has the clearest directional association with the unlabeled contrast at 40\% ($\rho=0.89$, exact permutation $p=0.035$, $n=6$). Activation low-rank energy is not monotone, and task-conditioned/effective-rank associations are mixed (the 30\% correlation is $\rho=0.60$, $p=0.24$). All descriptor--contrast pairs are reported below.

\begin{table*}[t]
\centering\scriptsize
\caption{Architecture-level response descriptors used in the Project1 alignment analysis. Values are parameter-weighted medians across prunable layers.}
\label{tab:mechanism-descriptors}
\begin{tabular}{l*{5}{r}}
\toprule
Model & Act. low-rank & Cancellation residual & Act. eff. rank & Grad. low-rank & Grad. eff. rank \\\\
\midrule
ResNet-50 & 0.083 & 0.953 & 0.122 & 0.020 & 0.477 \\\\
ConvNeXt-B & 0.126 & 0.958 & 0.019 & 0.030 & 0.077 \\\\
EfficientNet-B3 & 0.135 & 0.910 & 0.024 & 0.018 & 0.207 \\\\
MobileNetV2 & 0.119 & 0.936 & 0.035 & 0.014 & 0.479 \\\\
ViT-S/16 & 0.091 & 0.969 & 0.029 & 0.032 & 0.074 \\\\
Swin-T & 0.428 & 0.908 & 0.002 & 0.018 & 0.106 \\\\
\bottomrule
\end{tabular}
\end{table*}

\begin{table*}[t]
\centering\scriptsize
\caption{Spearman correlations between architecture descriptors and matched pruning contrasts. Exact permutation p-values use all 6! permutations; these are descriptive because $n=6$.}
\label{tab:mechanism-correlations}
\begin{tabular}{l*{4}{r}}
\toprule
Descriptor & Unlabeled 30\% & Unlabeled 40\% & Task-cond. 30\% & Task-cond. 40\% \\
\midrule
activation low-rank & -0.60 (0.243) & -0.49 (0.356) & -0.37 (0.498) & -0.14 (0.803) \\\\
cancellation residual & +0.77 (0.104) & +0.89 (0.035) & +0.09 (0.920) & -0.03 (1.000) \\\\
activation effective rank & +0.26 (0.659) & +0.14 (0.803) & +0.60 (0.243) & +0.37 (0.498) \\\\
gradient low-rank & +0.89 (0.035) & +0.77 (0.104) & -0.37 (0.498) & -0.26 (0.659) \\\\
readout alignment & -0.71 (0.137) & -0.77 (0.104) & +0.03 (1.000) & +0.26 (0.659) \\\\
gradient effective rank & -0.60 (0.243) & -0.54 (0.298) & +0.37 (0.498) & +0.14 (0.803) \\\\
\bottomrule
\end{tabular}
\end{table*}

\begin{figure}[t]
  \centering
  \includegraphics[width=0.99\linewidth]{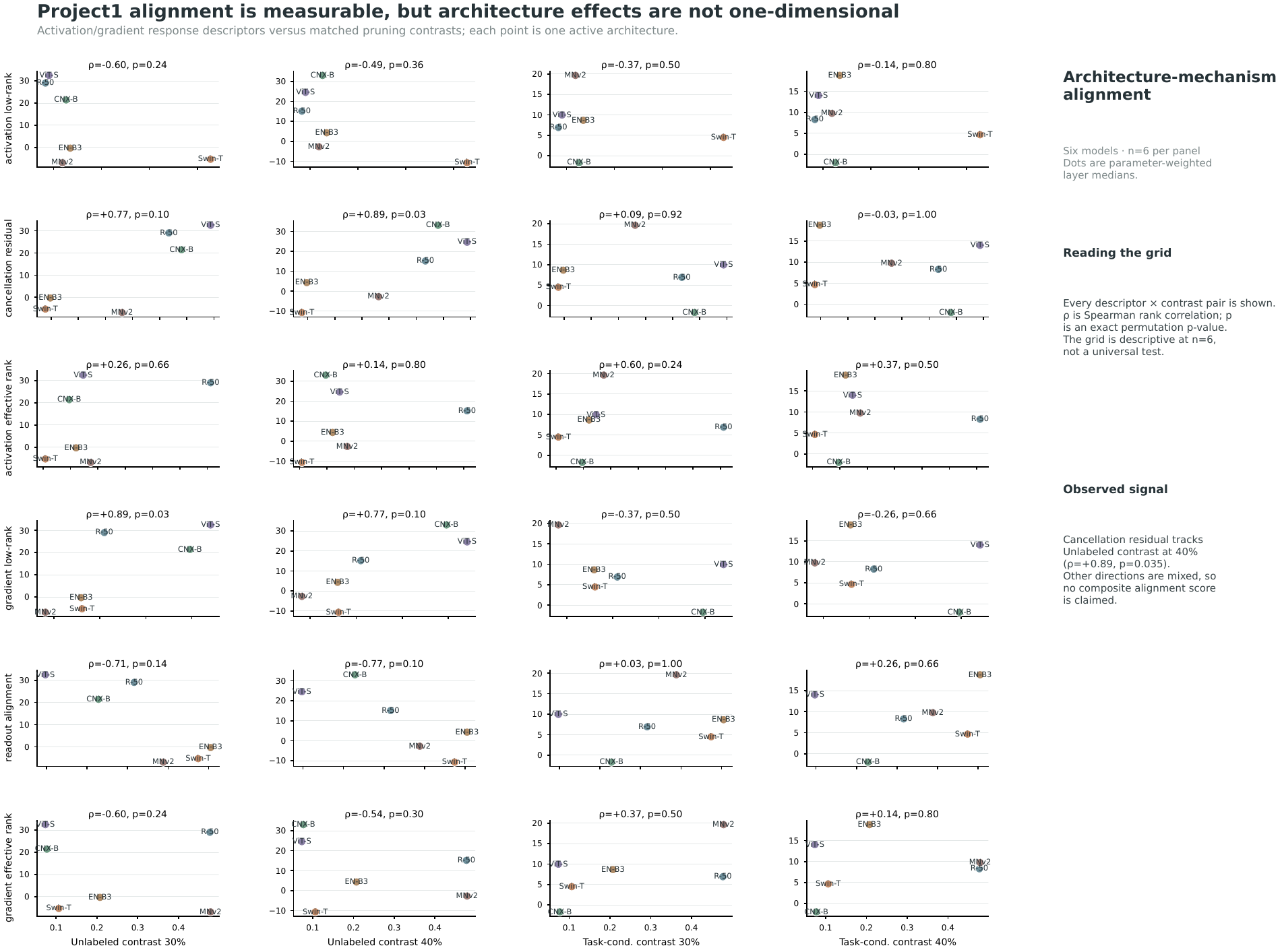}
  \caption{All descriptor--contrast pairs in the six-model mechanism test. Titles report Spearman $\rho$ and exact permutation $p$; every pair is shown to avoid selective emphasis.}
\end{figure}

\begin{figure}[t]
  \centering
  \includegraphics[width=0.95\linewidth]{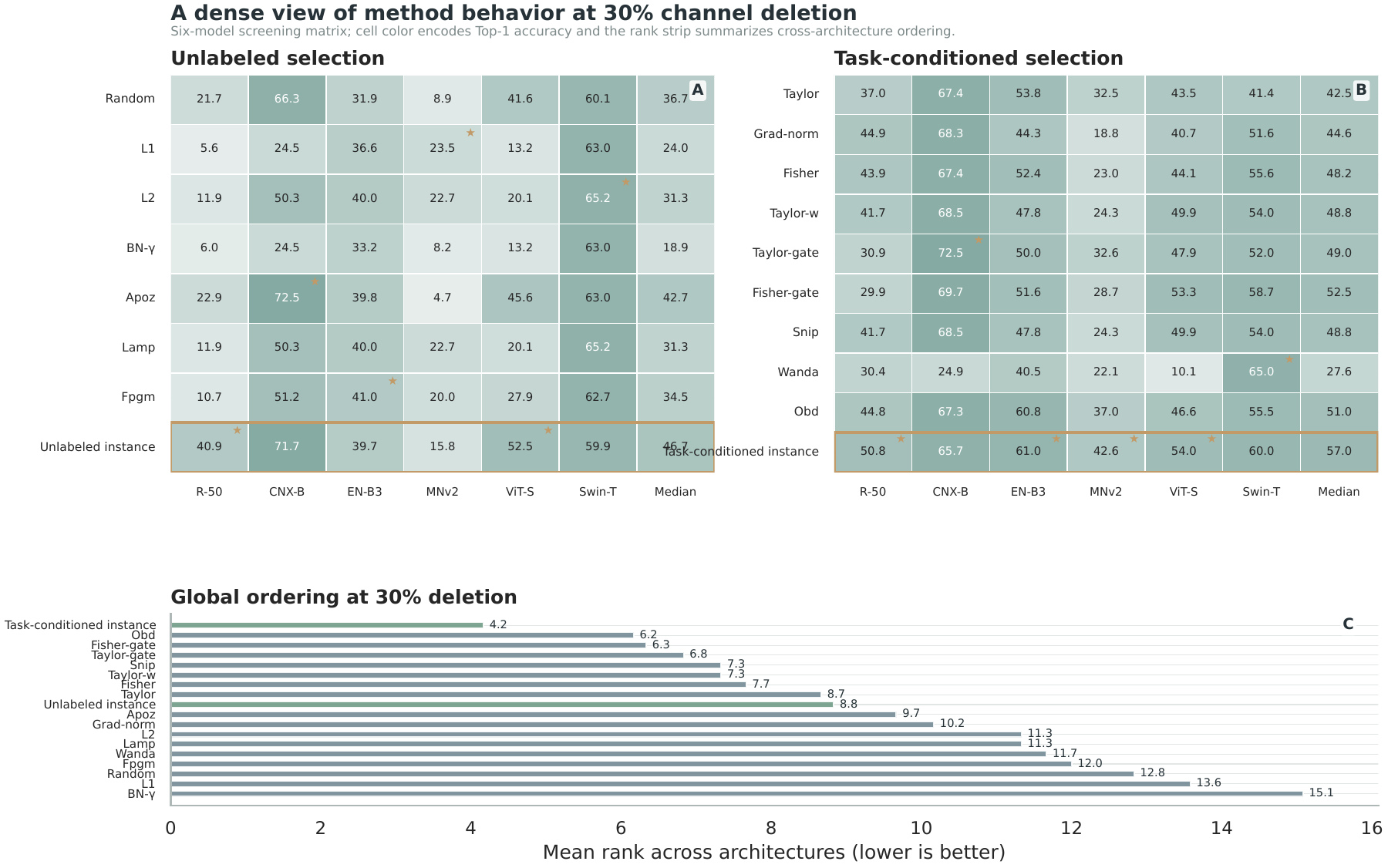}
  \caption{Complete configured methods at 30\% coordinate deletion for the six-model screen.}
\end{figure}
\begin{figure}[t]
  \centering
  \includegraphics[width=0.95\linewidth]{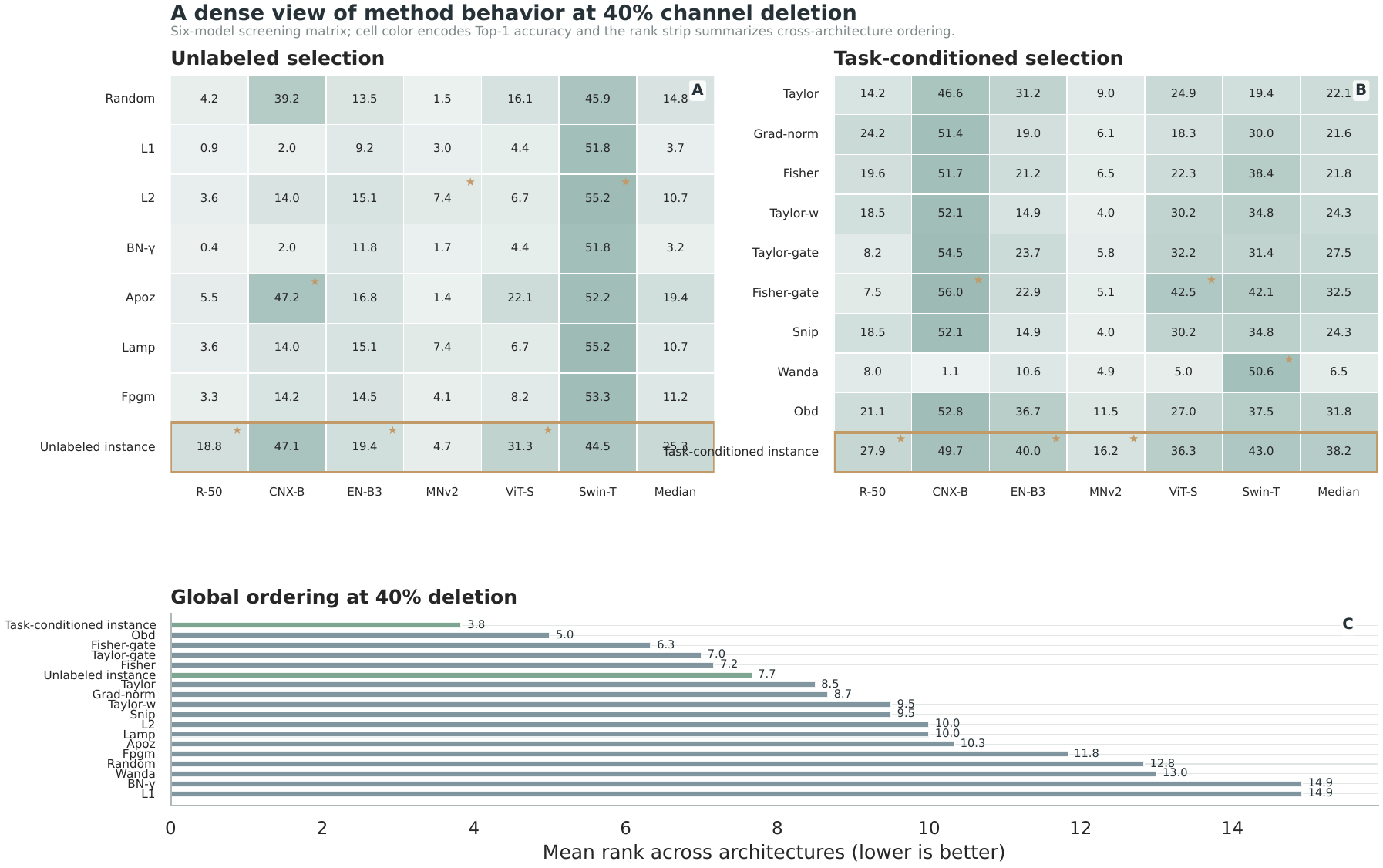}
  \caption{Complete configured methods at 40\% coordinate deletion for the six-model screen.}
\end{figure}
\begin{figure}[t]
  \centering
  \includegraphics[width=0.95\linewidth]{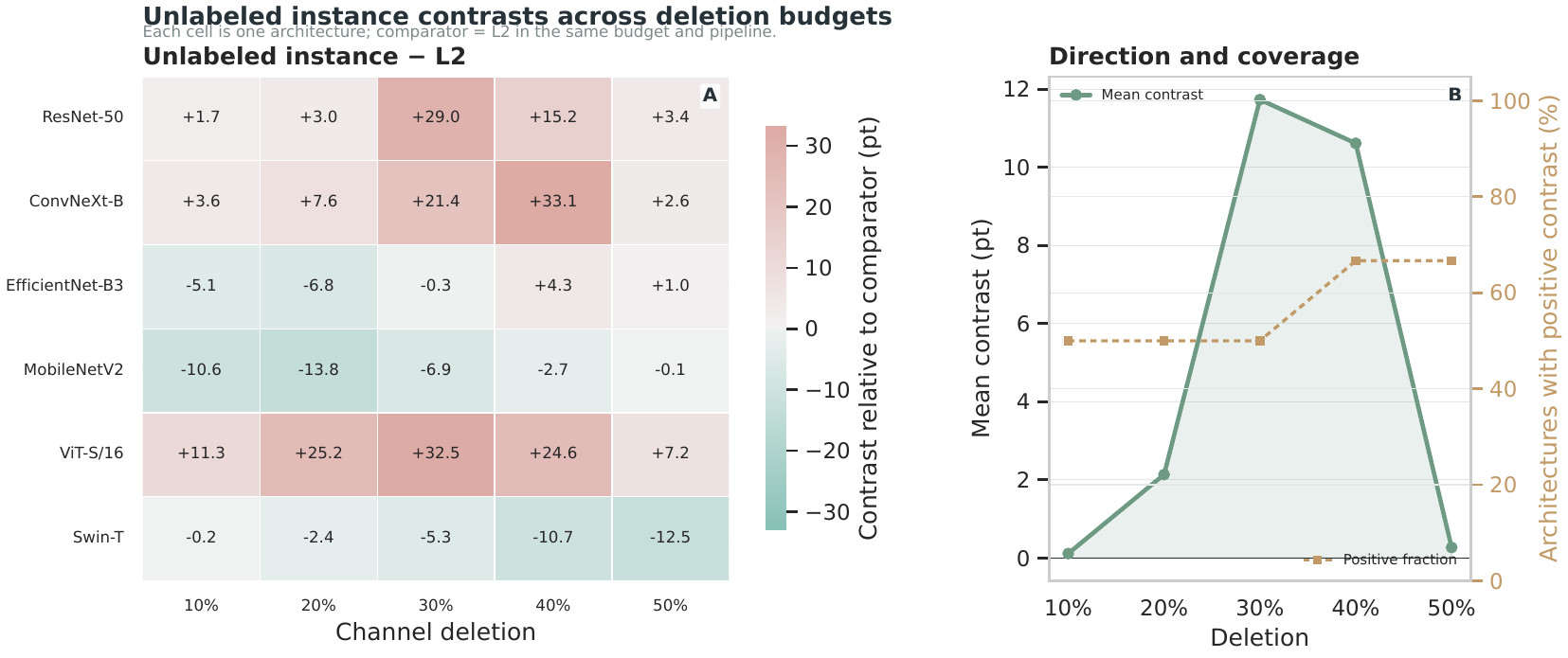}
  \caption{Unlabeled-instance Top-1 contrast relative to L2 across deletion rates and architectures. The right panel reports the mean contrast and the fraction of architectures with a positive contrast.}
\end{figure}
\begin{figure}[t]
  \centering
  \includegraphics[width=0.95\linewidth]{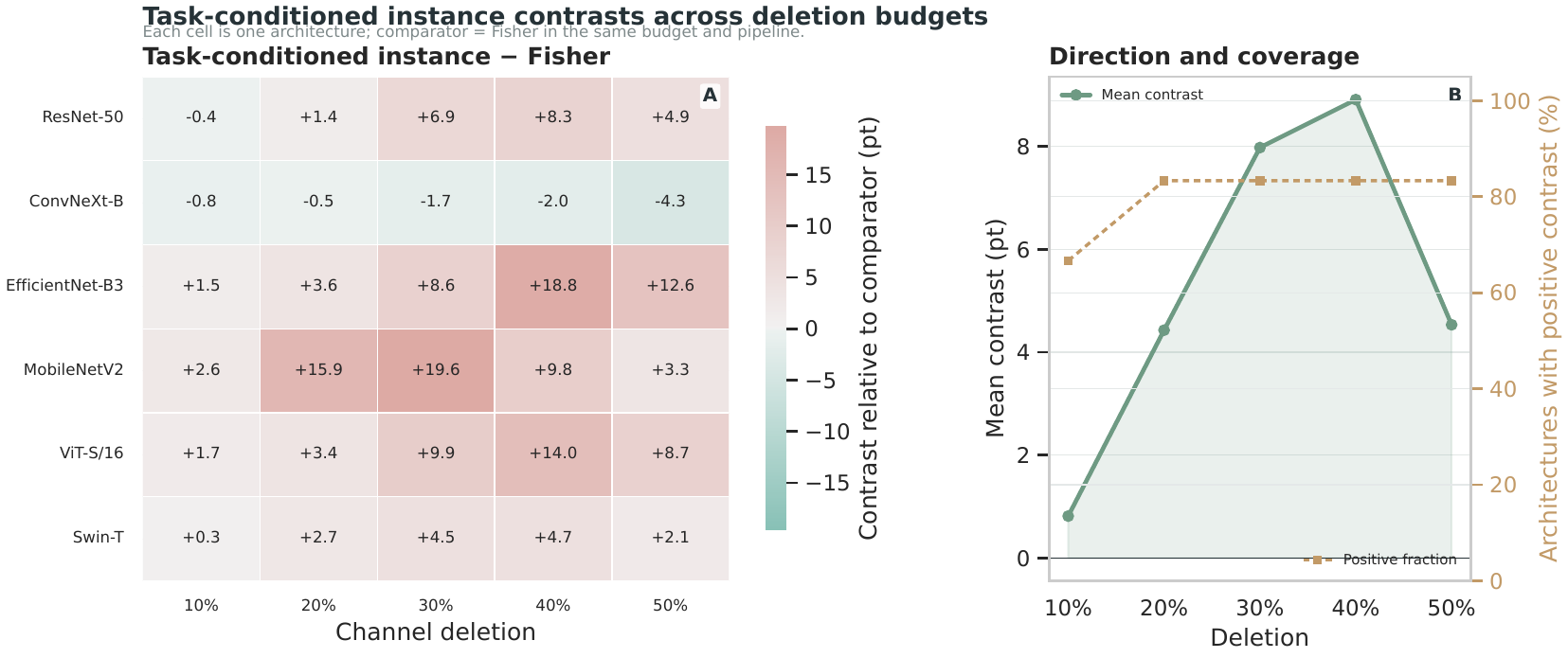}
  \caption{Task-conditioned Top-1 contrast relative to Fisher across deletion rates and architectures. The right panel reports the mean contrast and the fraction of architectures with a positive contrast.}
\end{figure}

The results show that response geometry changes selection outcomes under the fixed pruning and recovery pipeline, with architecture-dependent contrasts that motivate broader replicated studies.

\end{document}